\pdfoutput=1
\documentclass[11pt]{article}

\usepackage[]{ACL2023}

\usepackage{times}
\usepackage{latexsym}
\usepackage{graphicx}
\usepackage{multirow}
\usepackage{colortbl}
\usepackage{float}
\usepackage{graphicx} 
\usepackage{booktabs}
\usepackage{hyperref}

\usepackage[T1]{fontenc}
\usepackage[utf8]{inputenc}

\usepackage{microtype}

\usepackage{inconsolata}
\usepackage{float}

\title{AgriLLM:Harnessing Transformers for Framer Queries}

\author{Krish Didwania
\thanks{\ \ Equal contribution.}
  \thanks{\ \ Manipal Institute of Technology, Manipal Academy of Higher Education, Manipal, India}
  \thanks{\ \ Department of Computer Science and Engineering}\\
  \texttt{krishdidwania0674@gmail.com}
  \And
  Pratinav Seth
  \footnotemark[1]
  \footnotemark[2]
  \thanks{\ \ Department of Data Science \& Computer Applications}\\
  \texttt{seth.pratinav@gmail.com}
  \AND
  Aditya Kasliwal
  \footnotemark[2]
  \footnotemark[4] \\
  \texttt{kasliwaladitya17@gmail.com}
  \And
  Amit Agarwal
  \thanks{\ \ Enterprise Analytics \& Data Science,
  Artificial Intelligence - Center of Excellence,
  Wells Fargo International Solutions Private Limited,
  Bangalore, India}\\
  \texttt{amitagrawal1909@gmail.com}
}

\begin{document}
\maketitle
\begin{abstract}
Agriculture, vital for global sustenance, necessitates innovative solutions due to a lack of organized domain experts, particularly in developing countries where many farmers are impoverished and cannot afford expert consulting. Initiatives like Farmers Helpline play a crucial role in such countries, yet challenges such as high operational costs persist. 
Automating query resolution can alleviate the burden on traditional call centers, providing farmers with immediate and contextually relevant information.
The integration of Agriculture and Artificial Intelligence (AI) offers a transformative opportunity to empower farmers and bridge information gaps.
Language models like transformers, the rising stars of AI, possess remarkable language understanding capabilities, making them ideal for addressing information gaps in agriculture.
This work \footnote{Code available at:\url{https://github.com/krish0674/AgriLLM_Demo}} explores and demonstrates the transformative potential of Large Language Models (LLMs) in automating query resolution for agricultural farmers, leveraging their expertise in deciphering natural language and understanding context. Using a subset of a vast dataset of real-world farmer queries collected in India, our study focuses on approximately 4 million queries from the state of Tamil Nadu, spanning various sectors, seasonal crops, and query types.
\end{abstract}

\section{Introduction}
\begin{figure*}[]
  \centering
  \includegraphics[width=0.99\textwidth, height=0.27\textwidth]{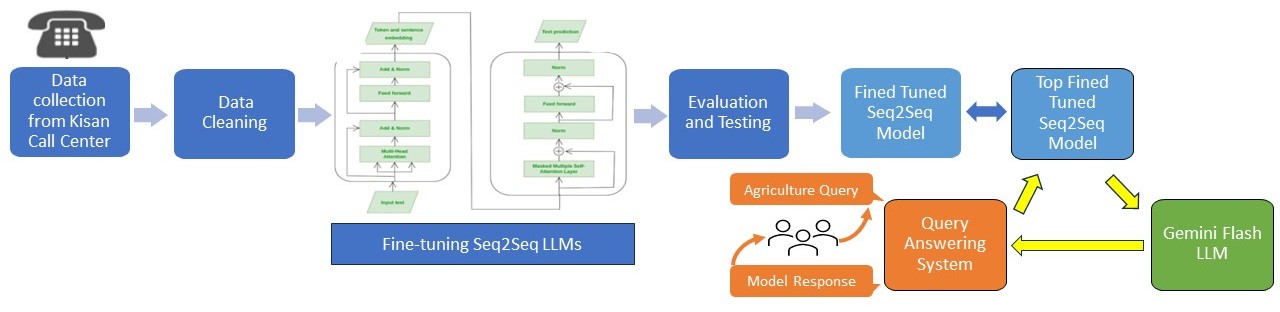}
  \caption{Flow Diagram of our methodology}
  \label{fig:method}  
\end{figure*}

Agriculture serves as a pivotal foundation for global economies, engaging a considerable workforce and making noteworthy contributions to the Gross Domestic Product (GDP) in numerous countries \cite{alston2014agriculture}.
This is especially true for developing countries, where agriculture often forms the backbone of the economy, providing employment for a large portion of the population and significantly driving economic growth and stability \cite{mosley2007aid}.

While playing a crucial role, the sector encounters numerous challenges, especially concerning the distribution of knowledge and accessibility to expert guidance.
These factors are essential for improving agricultural productivity and sustainability. Overcoming these challenges through better information dissemination and access to expert resources is critical for enhancing agricultural outcomes and ensuring long-term sustainability. By addressing these issues, developing countries can boost their agricultural productivity, foster economic development, and reduce poverty \cite{devlet2021modern}.

In recent years, the rapid digitization of the agricultural sector, facilitated by advancements in Information and Communication Technologies (ICT), has opened new avenues for addressing these challenges. 
Initiatives like the Kisan Call Center (KCC) in India have marked significant steps toward bridging the information gap faced by farmers \cite{zhang2016agricultural}. 
Nevertheless, limitations like network congestion, restricted operational hours, and the proficiency of call center agents emphasize the requirement for scalable and efficient solutions. 
To boost efficiency and effectiveness in the agricultural sector, there is an urgent demand for advanced solutions capable of addressing the various challenges robustly faced by farmers.

Conventional methods like manual surveys and on-site visits by experts, while beneficial, are frequently time-consuming and expensive \cite{jack2013market}. 
The irregular resource allocation makes it challenging to conduct these methods regularly, often taking several days or weeks. This underscores the need for an automated solution to promptly and efficiently address farmer queries.
Historically, automated query answering relied on keyword matching and predefined rules, leading to robotic and inaccurate responses. However, the recent emergence of Language Models and transformers signals a transformative phase. 

Recent progress in large language models (LLMs) and autoregressive LLMs such as ChatGPT, Gemini, Mistral, and Llama has demonstrated considerable promise. Through extensive pre-training, these models can produce high-quality, contextually relevant summaries that are both informative and accessible. However, they come with a substantial computational cost and often require fine-tuning for specific tasks, demanding significant computational resources and energy \cite{khemakhem2021causal}.

Sequence-to-sequence models present a promising solution, effectively addressing these computational challenges. Their encoder-decoder structure enables them to efficiently manage input and output sequences, making them well-suited for tasks with small input-output sizes. Compared to LLMs, they are more computationally efficient, requiring fewer resources for fine-tuning and inference while maintaining high performance and suiting some tasks better than causal models \cite{lei2018sequicity}.
These models also excel in scenarios where a specific tone, particularly one understandable to users in regions like developing countries, needs to be modeled. This is particularly relevant in regions where users are accustomed to a particular tone of English and generally prefer shorter, more keyword-highlighted conversations. This can be challenging for autoregressive LLMs when fine-tuned, as they are pre-trained on a large corpus of English text primarily contributed by people from America and Europe. This can lead to model forgetting information learned during pre-training, hallucinations, and subpar performance on downstream tasks.

Significant progress has been made in developing conversational AI interfaces tailored for various sectors, resulting in diverse solutions that meet the specific needs of stakeholders \cite{dwivedi2021artificial}. However, there has been limited exploration in the context of agricultural use cases. 
In particular, there has been no investigation into using Large Language Models (LLMs) for Natural Language Generation (NLG) to address query answering in the agricultural sector. This work demonstrates the NLG capabilities of LLMs in automating query resolution within this domain.

Leveraging their proficiency in interpreting natural language and understanding context, our study aims to present a reliable solution to the challenges faced by farmers in their daily lives. 
This research is pioneering in its application of LLMs for sequence-to-sequence generation in the agricultural context. 
Unlike previous efforts that did not fully utilize the advanced capabilities of LLMs, our approach harnesses these models for natural language generation using a subset of an extensive dataset of real-world farmer queries collected in India.
To enhance the quality and clarity of the answers, ensuring they are understandable to the minority of youths with basic education involved in agriculture, we integrated a pre-trained foundation LLM \cite{kivisild2003genetics}. This model enriches the structure and corrects the grammar of the answers generated by our base model.

\section{Related Works}
\subsection{Seq2Seq Transformers and LLMs}
Neural sequence-to-sequence (seq2seq) models have proven to be highly effective in handling various human language tasks and modeling sequential data. These models can process and generate human language, starting with the basics of natural language processing (NLP). Transformers, a significant advancement in NLP, enable more efficient handling of long-range dependencies in sequences. 

Seq2seq models, which build on transformers, encode an input sequence into a fixed representation and then decode it into an output sequence. Large language models (LLMs) and autoregressive models, such as GPT-3 and ChatGPT, further enhance the capabilities of seq2seq models by generating coherent and contextually appropriate text based on given prompts \cite{floridi2020gpt}. These sophisticated models offer a powerful approach for improving agricultural knowledge distribution and expert guidance accessibility by enabling advanced data processing and interpretation \cite{neubig2017neural}.

The evolution of Natural Language Processing (NLP) progressed from seq2seq models like recurrent neural networks to transformer models. Transformers revolutionized the field by enabling efficient parallel processing and capturing long-range dependencies in text with greater accuracy \cite{malte2019evolution}.
Historically, automated query answering relied on keyword matching and predefined rules, leading to robotic and inaccurate responses. However, the recent emergence of Language Models and transformers signals a transformative phase. 

Trained on extensive datasets of text and code, LLMs showcase exceptional proficiency in understanding and generating human-like language. 
This positions them as pioneers in automated query answering, providing unmatched fluency, dynamic comprehension, continuous learning, and expertise in multiple languages. 
LLMs handle intricate questions with nuance, decipher ambiguous queries, improve iteratively through user interactions, and surmount language barriers. 

Despite ongoing advancements, challenges like factual inconsistencies and biases necessitate attention. 
Nevertheless, the rapid evolution of LLM technology suggests a revolution in our interaction with information, making automated query answering more intuitive and advantageous.
\subsection{Agricultural Query Resolution}
Agriculture TalkBot \cite{vijayalakshmi2019agriculture} employed Natural Language Processing (NLP) based query retrieval alongside speech synthesis to create a more intuitive and accessible platform for addressing farmers' questions, narrowing the knowledge gap. 
AgriBot \cite{jain2019agribot} utilized the Kisan Call Center (KCC) and other datasets to provide guidance on weather conditions, market trends, and government policies using sentence embedding and Artificial Neural Networks for query retrieval-based answering. 
The LINE chatbot in Thailand demonstrated the practical application of these technologies in educating farmers about crop management, emphasizing the significance of chat-bots in the evolution of intelligent agricultural systems \cite{suebsombut2022chatbot}. 

Farmer-Bot \cite{darapaneni2022farmer} leveraging WhatsApp's popularity in India to provide a chatbot with RASA backend while utilizing the KCC dataset for the state of Assam, aiming to overcome limitations faced by traditional help centers. 
Farmer’s Friend \cite{venkata2022farmer} developed a similar multi-platform solution that aimed to help farmers through advanced NLP techniques. 
Recently, AgAsk \cite{koopman2023agask}, took a significant step by utilizing scientific literature to respond to intricate agricultural queries, highlighting the capabilities of information retrieval systems and setting a new standard for AI applications in agriculture.
\section{Methodology}
\subsection{Dataset Description}
In the realm of agriculture, timely and accurate information is crucial for the success and sustainability of farming practices \cite{vadivelu2013problems}. 
Farmers frequently encounter challenges that require immediate expert advice, ranging from pest management to crop rotation strategies \cite{RAO2007491}. 
Addressing this need for an efficient and accessible solution, the Indian Ministry of Agriculture and Farmers Welfare initiated the Kisan Call Center \cite{KCC-web-link} project, where farmers can call to ask queries which will be answered by domain experts.

The dataset comprises real queries posed by farmers and responses from domain experts. 
It encompasses essential details such as the farmer's location, crop type, nature of the query, and additional pertinent information about the specific crop in question.
This extensive dataset comprises more than 32 million queries, spanning from 2006 to 2023, covering all states in India. 
To align with our limited available computational resources, our research specifically concentrates on queries originating from the geographical region of Tamil Nadu.

\begin{table}[H]
\centering
\begin{tabular}{p{0.25\linewidth}|p{0.6\linewidth}}
\hline
\textit{Query} & {Fertilizer management for banana} \\ 
\textbf{KccAns} & spray borox 5g copper sulphate 5g zinc sulphate 5gmlit of water \\ 
\textbf{\textit{Preprocessed KccAns}} & spray borox 5 grams copper sulphate 5 grams zinc sulphate 5 grams per litre of water 
\\ 
\hline
\textit{Query} &{Basal application for paddy} \\ 

\textbf{KccAns} & apply DAP 50kg neemcake 10kg per ac \\ 

\textbf{\textit{Preprocessed KccAns}} & apply dap 50 kilograms neem cake 10 kilograms per acre \\ 
\hline
\end{tabular}
\caption{Preprocessing Examples}

\label{table:preprocessing}
\end{table}

\subsection{Data Preparation}
In this study, we work with the queries based on the geographic region of the state of Tamil Nadu, which contains 4 million records.
we created a split of 1\% each for testing and validation.
During the split, we ensured an equitable distribution of various query types and crop categories across all subsets. The data cleaning process required substantial effort to remove extraneous data and address observed inconsistencies.

Since the dataset consisted of transcripts of phone queries, it contained numerous run-on sentences, complicating the parsing and comprehension of the text. 
We thoroughly analyzed n-gram patterns to identify these blended phrases lacking clear boundaries. 
Due to the intricacy of the issue and the absence of an automated solution, manual correction became necessary. 
By carefully examining n-gram patterns, we manually separated these phrases in the samples, thereby enhancing the readability and structure of the dataset. 
The effects of modifications are detailed in Table \ref{table:preprocessing}.

During the data pre-processing phase, our main emphasis was on resolving abbreviations, ensuring case consistency, and handling data that combines numbers and text, with the numeric text retained due to its significance in providing context and detail to the textual data.
This thorough pre-processing significantly improved data quality and reliability, both qualitatively and quantitatively. 

Due to farmers' preference for their regional language and their limited English proficiency, with a vocabulary mostly consisting of certain keywords related to fertilizers, crops, and techniques, a large number of query and resolution instances in the dataset are grammatically incorrect. However, these queries are understandable to farmers and have a regional tone, which helps them confidently accept and understand the solutions, making them well-suited for the use case.

To improve the quality and readability of the answers, which should be understandable by the minority of youths involved who have basic education, we incorporated a pre-trained foundation model (i.e., Gemini Pro ). This model enriches and corrects the grammar of the answers produced by our base model, as shown in Figure \ref{fig:method}.

\begin{table}[H]
\centering
\begin{tabular}{p{0.3\linewidth}|p{0.6\linewidth}}
\hline
\textbf{Query} & \textbf{Generated Output} \\
\hline
asking about watermelon sowing season & recommended watermelon sowing season november - december \\
\hline
asking stem borer management brinjal & recommended for spray thiodicarb 2 grams per litre of water \\
\hline
leaf folder control paddy & recommended for spray cartaphydrochloride 2 grams per litre of water \\
\hline
paddy top dressing & apply urea 25 kilograms potash 15 kilograms micronutrient mixture 5 kilograms per acre \\
\hline
asking new farming introducing app & recommended for kisan suvidha app \\
\hline
\end{tabular}
\caption{Qualitative Results from fine-tuned Flan T5 Base}
\label{gentext1}
\end{table}

Table \ref{gentext1} shows some of the generated model outputs for the queries of the test set. Upon examination, these outputs align closely with the actual solutions to the problems as suggested by experts. This alignment indicates the model's effectiveness in generating accurate and relevant solutions, demonstrating its potential utility in real-world applications. 

Table \ref{ngentext} highlights cases where the model's outputs do not perfectly align with the actual answers in the test set. However, it is evident that the outputs are very close to the correct solutions, demonstrating the model's potential usefulness for users. 

\begin{table}[t]
\centering

\begin{tabular}{p{0.2\linewidth} | p{0.7\linewidth}}
\hline
\textit{Query:} & asking mulching sheet availability \\
\textbf{KccAns:} & recommended to contact Hosur Sri Vinayaka Trading Co phone - 08447518196 \\
\textbf{\textit{Generated Output:}} & recommended to contact Coimbatore Deputy Director of Horticulture phone - 0422-2453578 \\
\hline
\textit{Query:} & asking calcium deficiency tomato \\
\textbf{KccAns:} & recommended for spray calcium chloride 2 grams per litre \\
\textbf{\textit{Generated Output:}} & recommended for spray calcium chloride 5 grams per litre \\
\hline
\textit{Query:} & jamun thracnose management \\
\textbf{KccAns:} & recommended for spray Dithane 2 grams per litre of water \\
\textbf{\textit{Generated Output:}} & recommended for spray carbendazim 1 gram per litre of water \\
\hline
\end{tabular}
\caption{Drawbacks of the model compared to actual expert answers in a few cases}
\label{ngentext}
\end{table}

\subsection{Model Development}\label{sec:models} 

Our methodology can be observed in Figure \ref{fig:method}. We employed sequence-to-sequence modeling, specifically designed to transform small sequences of input data into corresponding sequences of output data, enabling the handling of intricate tasks like text translation, summarizing, and question-answering. 

The primary reason for using language sequence-to-sequence models for this task, rather than causal models, is their computational efficiency, resulting in faster query processing times. Additionally, given the nature of the datasets, farmers or users may require a clear, one-shot answer to their agricultural queries, which seq2seq transformers can effectively provide. 

To effectively prepare the model for the designated task, we conducted fine-tuning on various models such as BART \cite{lewis2019bart}, T5 \cite{raffel2023exploring}, and Pegasus \cite{pmlr-v119-zhang20ae}.
We also included indicBART \cite{Dabre_2022}  and a multilingual T5 model \cite{xue2021mt5} in our study, as they are capable of handling multiple languages. 
%The optimal model performance was noted with Flan T5 Base.

The fine-tuning process entailed configuring a batch size of 64 and a learning rate of $4 \times 10^{-5}$ with the AdamW \cite{loshchilov2019decoupled} optimizer and Cross Entropy Loss function \cite{mao2023crossentropy}. The finetuned models were evaluated through the metrics- BLEU score, ROUGE-1 score, and BERTscore \cite{zhang2020bertscore}, which comprises precision, recall, and F1 score as components of a reference-based metric.

\subsection{Enhancing Answer Tone and Grammar with Pre-Trained LLM}
As previously noted, the dataset contains a specific tone and may have grammatical errors since English is not the primary language of the local population in Tamil Nadu, and conversations may emphasize keywords. Consequently, the generated answers might not be easily understandable to a broader audience.

To address this challenge, we use a foundational auto-regressive large language model to improve the grammar and tone of the generated answers. This approach allows us to produce both a local version and an easily understandable version for an educated audience.

\begin{table}[t]
\centering

\begin{tabular}{p{0.2\linewidth} | p{0.7\linewidth}}
\hline
\textit{Query:} & paddy top dressing \\
\textbf{Generated Output:} &  apply urea 25 kilograms potash 15 kilograms micronutrient mixture 5 kilograms per acre\\
\textit{\textbf{Rephrased Output:}} &  Apply a fertilizer blend of 25 kilograms urea, 15 kilograms potash, and 5 kilograms micronutrient mixture per acre. \\
\hline
\textit{Query:} &  thrips control chilli
\\
\textbf{Generated Output:} & recommended for spray fipronil 2 millilitre per litreer of water \\
\textit{\textbf{Rephrased Output:}}  & The recommended dilution for fipronil is 2 milliliters per liter of water. \\
\hline
\textit{Query:} &  mealy bug management tuberose \\
\textbf{Generated Output:} &  recommended for spray profenophos 2 milli litre per acre ephate 1 grams litre of water\\
\textit{\textbf{Rephrased Output:}} &  The recommended application is 2 milliliters of profenophos spray per acre and 1 gram of ephate per liter of water.\\
\hline
\textit{Query:} &  top dressing fertilizer management tapioca
\\
\textbf{Generated Output:} & apply 35 kilograms urea 75 kilograms potashac 90 days after planting
 \\
\textit{\textbf{Rephrased Output:}} & Plants require 35 kg urea and 75 kg potash fertilizer 90 days after planting. \\
\hline
\end{tabular}
\caption{Enhancing the quality of generated answers through a foundation model for an educated audience.}
\label{gentext}
\end{table}
For this enhancement, we utilized Google's Gemini Flash model, which significantly improved the tone and grammatical accuracy of the generated answers. By employing this advanced tool, we produced coherent, professional, and beneficial responses for the end users, ultimately enhancing the overall user experience and the utility of our model's outputs.
The specific prompt used for this task was,
\begin{quote}
    Paraphrase and Correct Tone: <response>
\end{quote}
This prompt guided the model to effectively refine the language and structure of the seq2seq model's answers.

\begin{table}[H]
\centering
\small
\begin{tabular}{cccc}
\hline
                                                                                & \begin{tabular}[c]{@{}c@{}}Flesch\\ Kincaid\end{tabular} & \begin{tabular}[c]{@{}c@{}}Coleman\\ Liau\end{tabular} & \begin{tabular}[c]{@{}c@{}}Dale\\ Chall\end{tabular} \\ \hline
KccAns (Label)                                                                  & 7.865                                                    & \textbf{11.169}                                        & 14.894                                               \\
Flan T5 Base Output                                                             & 7.953                                                    & 11.578                                                 & 15.061                                               \\
\begin{tabular}[c]{@{}c@{}}Gemini Rephrased \\ Flan T5 Base Output\end{tabular} & \textbf{10.084}                                          & 11.5215                                                & \textbf{13.371}                                      \\ \hline
\end{tabular}
\caption{Readability Metrics for Text Outputs with Three Different Readability Scores on the Test Set. Here, KccAns refers to the actual answer given by the domain expert to the Query in the dataset.}
\label{tab:readability_metrics}
\end{table}

To evaluate the refined language we use three readability metrics: Flesch-Kincaid Grade Level (FKGL), Coleman-Liau Index (CLI), and Dale-Chall Readability Score (DCRS). 

FKGL estimates the U.S. school grade level needed to understand a text, with higher scores indicating more complex texts. CLI also estimates the grade level required for comprehension based on characters, words, and sentences, with higher scores signifying greater complexity. DCRS assesses readability based on familiar words, where lower scores mean the text is easier to read. 

In evaluating interpretability as shown in Table \ref{tab:readability_metrics}, rephrased outputs with higher FKGL scores suggest that the text is more detailed and sophisticated, potentially offering more nuanced and precise information, which can be beneficial in certain contexts. 

Additionally, a lower DCRS score in the rephrased output indicates that, despite the increased complexity, the text uses familiar words, making it more accessible and easier to understand. 

Thus, rephrased outputs showing higher FKGL and CLI scores and a lower DCRS score compared to the model's output demonstrate a balance of complexity and accessibility, enhancing readability and interpretability.

\begin{table}[H]
\centering
\scriptsize
\begin{tabular}{cccccc}
\hline
Model                                                                                   & Bl.            & Ro.            & P.             & R.             & F1             \\ \hline
\begin{tabular}[c]{@{}c@{}}T5-Small\\ \cite{raffel2023exploring}\end{tabular}           & 0.521          & 0.701          & 0.823          & 0.833          & 0.825          \\
\begin{tabular}[c]{@{}c@{}}BART-Base\\ \cite{lewis2019bart}\end{tabular}                & 0.55           & 0.72           & \textbf{0.836} & 0.837          & 0.835          \\
\begin{tabular}[c]{@{}c@{}}T5-Base\\ \cite{raffel2023exploring}\end{tabular}            & 0.547          & 0.718          & 0.831          & 0.843          & 0.834          \\
\begin{tabular}[c]{@{}c@{}}Flan-T5-Base\\ \cite{chung2022scaling}\end{tabular}          & \textbf{0.555} & \textbf{0.724} & 0.834          & \textbf{0.846} & \textbf{0.837} \\
\begin{tabular}[c]{@{}c@{}}Flan-T5-Small\\ \cite{chung2022scaling}\end{tabular}         & 0.53           & 0.708          & 0.824          & 0.836          & 0.828          \\
\begin{tabular}[c]{@{}c@{}}T5-Efficient-Small\\ \cite{raffel2023exploring}\end{tabular} & 0.522          & 0.701          & 0.822          & 0.832          & 0.824          \\
\begin{tabular}[c]{@{}c@{}}indicBART\\ \cite{Dabre_2022}\end{tabular}                   & 0.5            & 0.689          & 0.819          & 0.829          & 0.823          \\
\begin{tabular}[c]{@{}c@{}}mT5-Small\\ \cite{xue2021mt5}\end{tabular}                   & 0.542          & 0.715          & 0.83           & 0.84           & 0.833          \\
\begin{tabular}[c]{@{}c@{}}Pegasus-Xsum\\ \cite{pmlr-v119-zhang20ae}\end{tabular}       & 0.552          & 0.717          & 0.827          & 0.844          & 0.833          \\ \hline
\end{tabular}
\caption{Results of all Finetuned models on the Test Set. $Bl.$, $Ro.$, $P.$, $R.$, and $F_1$ stand for Blue Score, Rouge Score, Precision, Recall and Macro F1 scores (best results in {\bf bold}), exhibiting the superiority of \texttt{Flan T5 Base}.}
\label{tab:allmodelsres}

\end{table}

\begin{table*}[ht]
\centering
\small % Reduces font size to fit within two columns
\begin{tabular}{ccccccc}
\hline
Meta Data                   & Subset              & \begin{tabular}[c]{@{}c@{}}Bleu \\ Score\end{tabular} & Rouge1 & Precision & Recall & F1 Score \\ \hline
\multirow{2}{*}{Sector}     & Agriculture         & 0.559                                                 & 0.72   & 0.829     & 0.847  & 0.847    \\
                            & Horticulture        & 0.554                                                 & 0.716  & 0.827     & 0.841  & 0.832    \\ \hline
\multirow{3}{*}{Season}     & Rabi                & 0.566                                                 & 0.725  & 0.835     & 0.843  & 0.837    \\
                            & Kharif              & 0.549                                                 & 0.718  & 0.827     & 0.847  & 0.834    \\
                            & Jayad               & 0.539                                                 & 0.709  & 0.819     & 0.84   & 0.827    \\ \hline
\multirow{5}{*}{Query Type} & Plant Protection    & 0.5                                                   & 0.69   & 0.813     & 0.872  & 0.858    \\
                            & Nutrient Management & 0.602                                                 & 0.757  & 0.851     & 0.872  & 0.858    \\
                            & Fertilizer Use      & 0.628                                                 & 0.76   & 0.862     & 0.863  & 0.861    \\
                            & Cultural Practices  & 0.476                                                 & 0.679  & 0.792     & 0.828  & 0.806    \\
                            & Government Scheme   & 0.443                                                 & 0.649  & 0.73      & 0.824  & 0.767    \\ \hline
\end{tabular}
\caption{Results of Ablation studies with \texttt{Flan T5 Base} indicating Blue Score, Rouge Score, Precision, Recall, and Macro F1 scores for various meta-data subsets.}
\label{tab:subset}
\end{table*}

\section{Results and Analysis}

We performed experiments with various state-of-the-art and baseline model architectures as described in section \ref{sec:models}, and the results of fine-tuning are shown in Table \ref{tab:allmodelsres}. These quantitative results were obtained without using the Gemini API for rephrasing, underscoring the similarity in grammar and tone between the generated output and the expert answers provided in the dataset.

The Flan T5 Base \cite{chung2022scaling}, pegasus-xsum \cite{pmlr-v119-zhang20ae}, and Bart-base \cite{lewis2019bart} models exhibited the most impressive performance metrics, demonstrating superior text generation capabilities due to their architectural design and pretraining techniques when compared to other models.

To assess the consistency and robustness of the LLM models for the Query Generation Task across various distributions, we aimed to utilize metadata. This involved a detailed analysis of the models based on three key attributes: Seasons (Rabi, Kharif, and Jayad, which are local names for Indian agricultural seasons), Query Types (such as plant protection, nutrient management, fertilizer usage guidelines for specific conditions, government schemes, and regional cultural practices), and Crop Sectors (e.g., horticulture crops like fruits and vegetables, and other agricultural crops). Table \ref{tab:subset} presents the performance analysis of our top fine-tuned model, Flan T5 Base, demonstrating well-rounded results. The model exhibits consistently strong performance across different sectors and seasons, though some variability is observed among query types. Notably, queries related to fertilizer usage and nutrient management yielded the best outcomes.

We saw consistent performance trends across different attributes in all the models we developed, though with increased standard variation. Thus, we can conclude that its capabilities are robust and consistent across almost all diverse scenarios.

\section{Conclusion and Future Works}
In this work, we harness the power of language models for sequence-to-sequence text generation to aid in addressing farmer queries over the dataset, which consists of authentic farmer query conversations with domain experts from the KCC dataset. 
Our approach involved rigorous data cleaning and preprocessing to eliminate noise from the dataset, marking one of the initial applications of LLMs for Natural Language Generation in this domain. 
The fine-tuned models demonstrate robust performance and effective generalization across various attributes, resulting in high-quality text generation. 

By enhancing tone and grammar with pre-trained large language models (LLMs), the rephrased output can be tailored to a more general and wider educated audience rather than just the local population. This approach has the potential to contribute to societal well-being by alleviating the workload on conventional call centers and domain experts, providing farmers with immediate and contextually relevant information.

This, in turn, establishes the groundwork for a more inclusive and responsive agricultural ecosystem, ensuring universal access to vital knowledge for all farmers and fostering a sustainable and prosperous future for global agriculture.

In future research endeavors, we aim to enhance our proposed methodology by integrating metadata collected alongside farmer inquiries to improve the training using Large Language Models (LLMs), enhancing their effectiveness. 

Furthermore, we plan to expand our work to diverse regions across India present in the KCC Dataset, enabling a more comprehensive resolution of farmers' concerns and collaborating with regional languages relevant to offer more localized assistance.

\section{Ethical and Societal Implications}
We examine the use of large language models (LLMs) to automate responses to farmer queries in India in this work. Given the crucial role agriculture plays in India's economy and the livelihoods of a significant portion of the population, it is vital to consider the ethical and societal implications of deploying such technologies. We recognize the potential benefits, such as enhancing farmers' access to timely and accurate information, which can improve crop management, increase productivity, and ultimately strengthen food security. However, it is equally important to carefully define what constitutes "positive impact" in this context.

A crucial consideration is the accessibility of these technologies for all farmers, including those in remote or underserved areas. The deployment of LLMs must be inclusive, ensuring it does not worsen existing inequalities or exclude specific groups. Integrating this technology with accessible services, such as phone-based systems, could be particularly beneficial. Additionally, the accuracy and reliability of LLM-generated information are paramount, as incorrect recommendations could lead to adverse outcomes for farmers. 

The use of LLMs in addressing farmer queries also raises complex issues of accountability and liability. It is essential to determine who is responsible if AI-generated advice results in crop failures or financial losses and to establish fair systems of redress. Therefore, a trial of such systems should be conducted to compare their actual effectiveness with user satisfaction. An expanded study incorporating multimodal features, such as location, crop type, and season, could improve the model and improve outcomes. Furthermore, this technology may drive long-term societal changes, potentially reshaping farmer-consumer relationships and rural-urban dynamics as agriculture becomes more data-driven and efficient.

We must also consider how this technology interacts with traditional knowledge systems. There is a risk that AI-driven advice could inadvertently undermine local, traditional farming wisdom passed down through generations, leading to a homogenization of agricultural practices and a reduction in crop diversity and resilience. Balancing technological progress with the preservation of cultural heritage and biodiversity will be a key challenge.

In conclusion, while automating farmer queries through LLMs offers promising opportunities to support India's agricultural sector, it is essential that these efforts are guided by a strong ethical framework. We advocate for a broad discussion on what constitutes "positive impact" in this context, ensuring that the benefits of such technologies are equitably distributed and contribute to the overall well-being of the farming community.
% \section{Acknowledgments}
% We would like to thank Mars Rover Manipal, an interdisciplinary student project team of MAHE, for providing the necessary resources for our research. 

\bibliography{main-tex}
\bibliographystyle{acl_natbib}

\end{document}